\documentclass[default,iicol]{sn-jnl}

\usepackage[x11names, svgnames, rgb]{xcolor}
\usepackage[utf8]{inputenc}
\usepackage{tikz}
\usetikzlibrary{snakes,arrows,shapes}
\usepackage{amsmath}
\usepackage{amsfonts}
\usepackage{amssymb}

\usepackage{graphicx}%
\usepackage{multirow}%
\usepackage{amsmath,amssymb,amsfonts}%
\usepackage{amsthm}%
\usepackage{mathrsfs}%
\usepackage[title]{appendix}%
\usepackage{xcolor}%
\usepackage{textcomp}%
\usepackage{manyfoot}%
\usepackage{booktabs}%
\usepackage{algorithm}%
\usepackage{algorithmicx}%
\usepackage{algpseudocode}%
\usepackage{listings}%
\usepackage{natbib}%

\theoremstyle{thmstyleone}%
\theoremstyle{thmstyletwo}%

\theoremstyle{thmstylethree}%

\begin{document}

\title[Article Title]{Spectral Normalization and Dual Contrastive Regularization for Image-to-Image Translation}


\author[]{\fnm{Chen} \sur{Zhao}}\email{2518628273@qq.com}

\author*[]{\fnm{Wei-Ling} \sur{Cai}}\email{caiwl@njnu.edu.cn}

\author[]{\fnm{Zheng} \sur{Yuan}}\email{l1455341238@qq.com}

\affil[]{\orgdiv{Department of Computer Science and Technology}, \orgname{Nanjing Normal University}, \orgaddress{ \city{NanJing},  \country{China}}}





\abstract{Existing image-to-image (I2I) translation methods achieve state-of-the-art performance by incorporating the patch-wise contrastive learning into Generative Adversarial Networks. However, patch-wise contrastive learning only focuses on the local content similarity but neglects the global structure constraint, which affects the quality of the generated images. In this paper, we propose a new unpaired I2I translation framework based on dual contrastive regularization and spectral normalization, namely SN-DCR. To maintain consistency of the global structure and texture, we design the dual contrastive regularization using different deep feature spaces respectively. In order to improve the global structure information of the generated images, we formulate a semantic contrastive loss to make the global semantic structure of the generated images similar to the real images from the target domain in the semantic feature space. We use Gram Matrices to extract the style of texture from images. Similarly, we design a style contrastive loss to improve the global texture information of the generated images. Moreover, to enhance the stability of the model, we employ the spectral normalized convolutional network in the design of our generator. We conduct comprehensive experiments to evaluate the effectiveness of SN-DCR, and the results prove that our method achieves SOTA in multiple tasks. The code and pretrained models are available at \href{https://github.com/zhihefang/SN-DCR/tree/main/SN-DCR-main}{https://github.com/zhihefang/SN-DCR}.}




\maketitle

\section{Introduction}\label{sec1}

Image-to-image translation (I2I) tasks aim to map an input image from the source domain into the target domain while retaining its original content and structure. In many I2I tasks, it is impossible to collect paired training data, and therefore adversarial loss \cite{Goodfellow} suffers from the collapse of the content and structure. In order to alleviate this problem, many typical works \cite{Benson1992} use cycle consistency loss \cite{Benson1992}, which enforces the consistency between the input images and the reconstructed images by an inverse mapping of the generated image. Motivated by vision transformer (ViT), the recent cycle consistency based work \cite{torbunov2023uvcgan} explored the generator with a ViT for unpaired image-to-image translation. However, cycle consistency requires that the mapping between the two domains be a bijection, which is too restrictive \cite{Dukowicz1984}.

\begin{figure*}[t]
	\centering
	\includegraphics[width=1.0\linewidth]{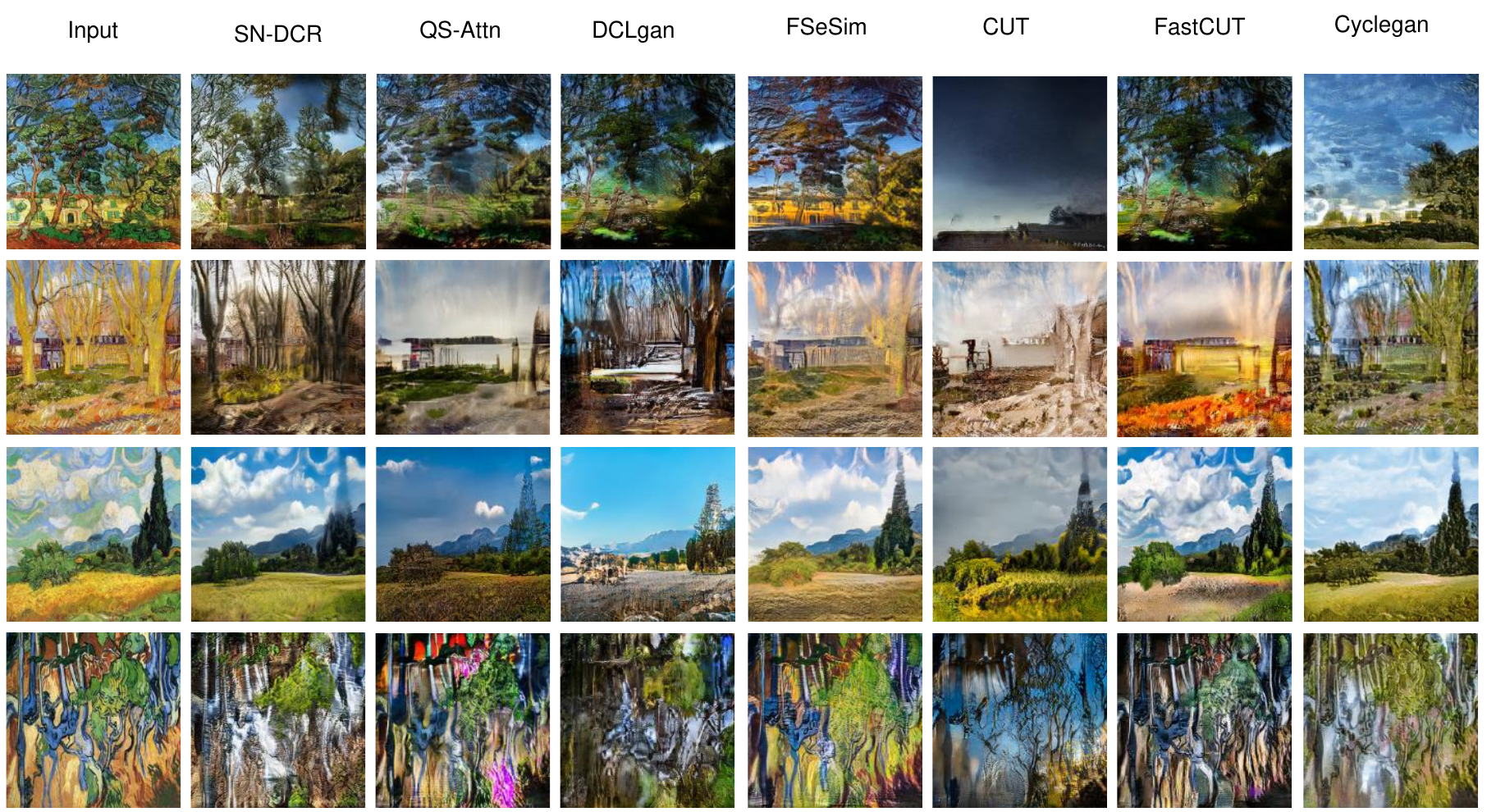}
	\caption{
		Visual comparison results with all baselines on the Van Gogh$\rightarrow$Photo dataset. It is obvious that the global structure of the images generated by previous methods is corrupted, while our SN-DCR is able to preserve the global structure information and generate the photos with more natural details. Our SN-DCR performs better in terms of global structure and texture. Note that CUT and Cyclegan fail to  generate a valid output at the first input.
	}
	\label{fig:framework}
	 \vspace{10pt}
	
\end{figure*}

Recently, inspired by the success of contrastive learning, CUT \cite{Dukowicz1984}  proposed patch-wise contrastive learning to maximize the mutual information between the same location of input and generated images, which introduced the contrastive learning into I2I translation for the first time. DCLGAN \cite{Margolin2003} proposed a method based on patch-wise contrastive learning and dual learning settings (using two generators and two discriminators). Although DCLGAN improves the quality of the generated images, an additional generator and discriminator need to be trained, which increases the training costs. F-LSeSim \cite{Chuanxia} utilized patch-wise contrastive learning by computing the learned self-similarity. However, it relies on VGG features to measure similarity, which reduces training efficiency. In previous studies, the query for contrastive learning was selected randomly from the generated images, which is an obvious problem since some locations contain less information from the source domain. Therefore, QS-Attn \cite{Xueqi} designed a query-selected attention module by intentionally choosing significant anchors for patch-wise contrastive learning. Recently, a novel I2I method \cite{song2023shunit}  based on style harmonization was proposed, which leverages two distinct styles: class-aware memory style and image-specifc component style. Gou et al. \cite{gou2023multi} proposed multi-feature contrastive learning (MCL) to construct a patch-wise contrastive loss using the feature information of the discriminator output layer for I2I tasks. 

The previous works retain content consistency of the generated images via patch-wise contrastive learning without any regularization. However, only patch-wise contrastive learning is unable to effectively maintain the overall structure and texture of the images, as it only focuses on the local content similarity but neglects the global structure constraint. This issue affects the quality of the generated image.

In this paper, we propose a new I2I translation framework based on dual contrastive regularization and spectral normalization (SN-DCR). To achieve a global constraint of structure and texture in an unpaired manner, we formulate two new global contrastive loss functions to supplement the patch-wise contrastive loss, called dual contrastive regularization (DCR). DCR contains two parts: one is the semantic contrastive loss and the other is the style contrastive loss. Specifically, the semantic contrastive loss is proposed to improve the global structure information of the generated images, which encourages the generated images and the real images of the target domain (positives) to pull together in the semantic feature space while pushing the generated images away from the real images of the source domain (negatives). For example, the semantic structure information of the generated dog should be similar to the real dog in the semantic feature space. The style contrastive loss is developed to achieve the global texture consistency, in which the feature space capturing texture maps \cite{Leon} is adopted to represent the style of texture. 

Furthermore, it is well known that the training of GAN is unstable, and there are some issues such as mode collapse and convergence difficulties. To alleviate these issues, we employ the spectral normalization  \cite{Kucharik2012} in the design of our model, which enhances the stability of training. Moreover, to further boost the feature representation ability of our model, Frequency Channel Attention Network (FCANet) \cite{Zequn} is introduced to improve the translation performance. We demonstrate that our designed generator is effective through ablation experiments. In order to perform a better patch-wise contrastive loss, the qs-attn module is introduced in this paper. We conduct the comprehensive experiments to evaluate the effectiveness of SN-DCR, and the results prove that our method achieves SOTA in multiple tasks.

In summary, the main contributions of our method are three folds:
\begin{itemize}
\vspace{10pt}
\item We propose a novel unpaired I2I translation framework via dual contrastive learning and spectral normalization, namely SN-DCR. Our proposed SN-DCR focuses not only on local content similarity but also on the consistency of the global structure and style, enabling the generation of high quality images with more natural structure and texture information. 
\vspace{10pt}
\item	To improve the global information of the generated images, we design dual contrastive regularization (DCR), which can achieve the consistency of the global structure and texture. Our proposed DCR can be viewed as a universal regularization to enhance quality of the generated images.
\vspace{10pt}
\item	Experimental results compared with SOTAs clearly demonstrate that our SN-DCR exhibits superiority over prior unsupervised I2I translation approaches. Furthermore, we conduct comprehensive ablation experiments to scrutinize each of our contributions, and prove the effectiveness of each element. 
\vspace{10pt}

\end{itemize}

\section{ RELATED WORK}\label{sec2}
\vspace{-3pt}
\subsection{Image-to-Image Translation}

GANs \cite{Goodfellow} \cite{cGAN} \cite{sgan} have obtained great success, especially in I2I translation and the key idea is adversarial loss \cite{Goodfellow}. I2I translation can be categorized into two groups: a paired setting \cite{Phillip} \cite{Liska2010} \cite{Taesung} (supervised) and an unpaired setting (unsupervised)\cite{Benson1992}\cite{Zili}\cite{Huang}\cite{Sagie} . Paired setting means that each image from the  source domain has a corresponding label, which can be regarded as classic GAN. In order to further improve the quality of the generated images, SPADE introduces a the spatially-adaptive normalization layer. However, it is difficult to obtain paired training data, as a result, current methods \cite{Benson1992} are usually based on unpaired settings, which are developed based on one assumption: cycle-consistency. For example, CycleGAN \cite{Benson1992}, Dual-GAN \cite{Zili} and MUNIT \cite{Huang} train cross-domain GANs with cycle-consistency loss. CycleGAN learns two mappings simultaneously via translating an image to the target domain and back preserving the fidelity of the input and the reconstructed image. However, the assumption of cycle consistency that two domains can be mapped in both directions is too strict to obtain sufficient context. To alleviate the issue, many methods have tried to break the cycle consistency. DistanceGAN \cite{Sagie} proposes a distance constraint that allows unsupervised domain mapping to be one-sided. GC-GAN \cite{THuan} enforces geometry consistency as a constraint for unsupervised domain mapping. CUT \cite{Dukowicz1984} introduces patch-wise contrastive learning into I2I translation, which significantly improves the quality of translation. However, only patch-wise contrastive loss is unable to effectively maintain the global structure and texture of the generated images.

\vspace{-5pt}
\subsection{Contrastive Learning}
\vspace{-3pt}
Recent studies of the self-supervised learning  \cite{Kaiming} \cite{Ting} \cite{sung2018learning} \cite{zhang2020reinforced} show its strong ability to represent an image without labels, particularly with the help of the contrastive loss \cite{Mathilde} \cite{Xinlei}. Its idea is to perform the instance-level discrimination and learn the feature embedding, by pulling the features from the same image together and pushing those from different ones away. It has been observed that contrastive loss has been utilized in several low-level visual tasks, resulting in outstanding performance in various applications such as style transfer \cite{zhang2023csast}, image generation \cite{phaphuangwittayakul2023few}, image smoothing \cite{zhu2023structure}, image super-resolution and Deraining \cite{wu2023practical, chang2023unsupervised}. PatchNCE \cite{Dukowicz1984} proposes patch-based contrastive learning, which uses a noise-contrastive estimation framework by learning the correspondence between the patches of the input image and the corresponding generated image patches.  Excellent results are achieved and the recent methods \cite{Margolin2003} \cite{Chuanxia} \cite{Xueqi} \cite{Weilun} \cite{chen} also obtained better performance by utilizing the idea of patch-wise contrastive learning. DCLGAN \cite{Margolin2003} proposes a method based on patch-wise contrastive learning and dual learning settings (using two generators and two discriminators). Although DCLGAN improves the quality of the generated images, an additional generator and discriminator need to be trained, which increases the training costs. QS-Attn \cite{Xueqi} designed a query-selected attention module by intentionally choosing significant anchors for patch-wise contrastive learning. MCL \cite{gou2023multi} designed multi-feature contrastive learning (MCL) to construct a patch-wise contrastive loss using the feature information of the discriminator output layer. In parallel to these various designed methods, we mainly explore global regularization by introducing the idea of global contrastive loss. Therefore, we propose a new unpaired I2I translation framework based on dual contrastive regularization and spectral normalization to improve the global information of the generated images, which consistently shows better results.

\begin{figure*}[t]
	\centering
	\includegraphics[width=1.0\linewidth]{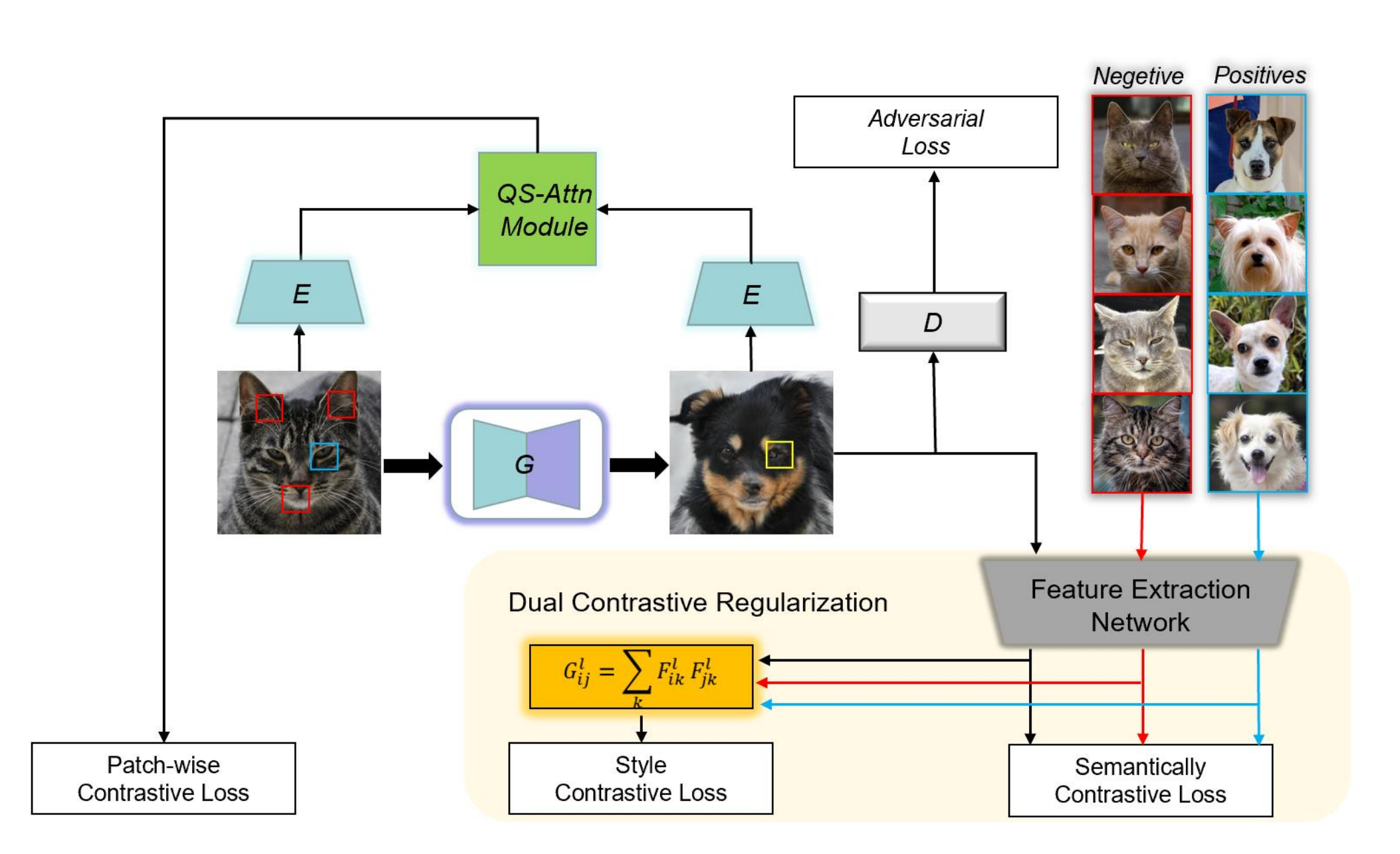}
	\caption{
		Overall framework of our proposed SN-DCR. A cat (the input image) is translated by the generator G into a dog(the generated image).  We introduce a dual contrastive regularization that combines both semantic and style contrastive loss to effectively pull the generated image closer to the real images of the target domain (positives) while pushing it away from the real images of the source domain (negatives).
	}
	\label{fig:framework}
\end{figure*}

\vspace{-3pt}
\section{OUR METHOD}\label{sec3}
\vspace{-5pt}
\subsection{Overall Framework}
Given an input image $\mathrm{~x~}\in\mathbb{~R}^{H\times W\times3}$ from the source domain X, our goal is to translate it into G(x) in the target domain Y via the adversarial loss, having no apparent difference with the real image$\mathrm{~y~}\in\mathbb{~R}^{H\times W\times3}$ from the domain Y. The framework of SN-DCR is shown in Fig.2, which consists of a generator G and a discriminator D. We divide the generator into two parts, the first part is defined as the encoder E, and the other part is defined as the decoder. SN-DCR includes three loss functions, the adversarial loss , patch-wise contrastive loss and dual contrastive regularization. In order to perform better patch-wise contrastive loss, we introduce the qs-attn module (global) in our proposed framework.

We expect the generated image G(x) to be as similar as possible to the real image $\mathrm{~y~}\in\mathbb{~R}^{H\times W\times3}$ from the domain Y , and the generator G can be mapped from the domain X to the domain Y through the adversarial loss. The adversarial loss is as follows:

\begin{eqnarray}
	\mathcal{L}_{adv}\!=\!\mathbb{E}_{y\in Y}\!\log\! D(\mathbf{y})\!+\!\mathbb{E}_{x\in X}\log(1\!-D(G(\mathbf{x}))),
\end{eqnarray}

\begin{figure*}[t]
	\centering
	\includegraphics[width=1.0\linewidth]{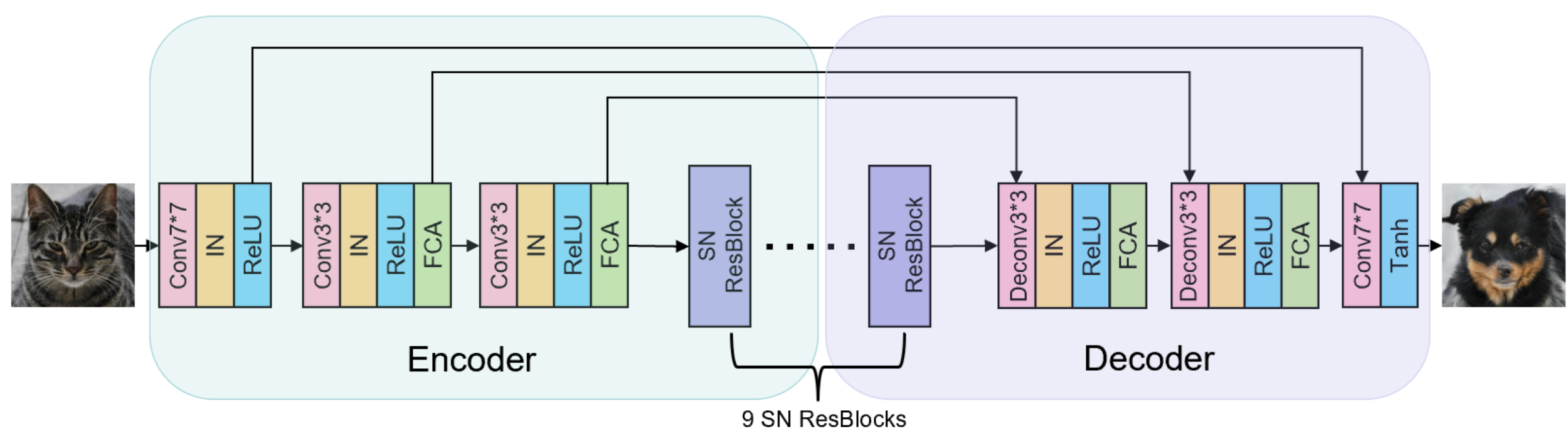}
	\caption{
		Our proposed spectral normalized generator, denoted as G, incorporates the use of InsNorm (IN) and a Frequency Channel Attention Network (FCA). Additionally, it utilizes our novel spectral normalized residual block (SN ResBlock), with nine such blocks present in the middle of the architecture. In line with the configuration of CUT, we extract features from five different layers to compute the multi-layer patch-wise contrastive loss. These layers include RGB pixels, the initial two downsampling convolutions, and the first and fifth residual blocks.
	}
	\label{fig:framework}
\end{figure*}

\subsection{Network Architecture}

\noindent\textbf{Generator.} We employ a UNet-based network with 9 residual blocks as the generator module. As known, the training process of GANs is very unstable, and problems such as mode collapse and convergence difficulties often occur. We employ SNconv in the design of residual blocks, which enhances the stability of training. In addition, to further boost the performance of our proposed SN-DCR, an up-to-date attention mechanism Frequency Channel Attention Network(FCANet) is introduced into our network.
As illustrated in Fig. 3, given an input image x, the generator G can map x to G(x) in the target domain. To achieve this goal, G is supposed to preserve both image structures and details when translating. Motivated by previous studies, we exploit an encoder-decoder network with nine residual blocks as the generator. Given a cat, we first employ an initial layer and two down-sampling layers to encode the input image into a low-resolution feature map. Then, nine SN ResBlocks are adopted to extract more complex and deeper features in the low-resolution space. Fig. 4 shows the detailed structure of SN ResBlock. After that, we employ the corresponding two up-sampling layers and a 7 * 7 convolutional layer to output a dog. Moreover, as mentioned above, we introduce FCANet in the generator design to further enhance the ability of SN-DCR. FCANet combines the channel attention mechanism with the discrete cosine transform cleverly, and expands on the basis of SENet \cite{JieHu} to obtain a new multi spectral channel attention mechanism. FCANet enables our model to learn the weights from different feature maps adaptively. Moreover, after the introduction of FCANet, the results of ablation study indicate that the performance of the proposed model can be improved significantly.

\noindent\textbf{Discriminator.}  We use the same PatchGAN discriminator \cite{patchGAN} architecture as CycleGAN and Pix2Pix which uses local patches of sizes 70x70 and assigns every patch a result. This is equivalent to manually crop one image into 70x70 overlapping patches, run a regular discriminator over each patch, and average the results. For instance, the discriminator takes an image from either domain X or domain Y , passes it through five downsampling Convolutional Normalization LeakeyReLU layers, and outputs a result matrix of 30x30. Each element corresponds to the classification result of one patch. Following CycleGAN  and Pix2Pix, in order to improve the stability of adversarial training, we use a buffer to store 50 previously generated images.

\begin{figure}[t]
	\centerline{\includegraphics[scale=0.63]{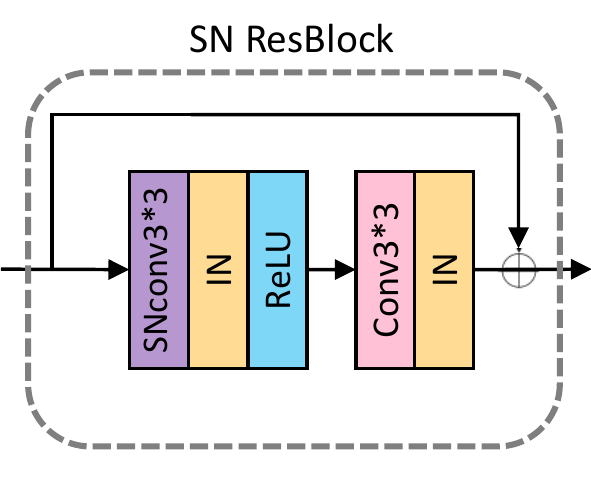}}
	\caption{SN Residual block. The SN residual block can enhance the stability of training and assist the generator to extract more complex features.\label{fig1}}
	\vspace{-5pt}
\end{figure}

\subsection{Dual Contrastive Regularization}
\vspace{-2pt}
We adopt the real images of the target domain and source domain as the positives and negatives to improve better quality of the generated images. DCR constrains the generator by two different feature spaces. Note that, to ensure the flexibility of our proposed method, these positives and negatives are randomly chosen. To achieve a constraint of the global structure, we propose a semantic contrastive loss, which aims to encourage the generated image G(x) to be close to the positives P while keeping away from the negatives N. For the feature space, inspired by AECR-Net \cite{Haiyan} ,we employ a pre-trained VGG-16 network to extract the feature maps $f\in\mathbb{R}^{C\times H\times W}$. 

The semantic contrastive loss can be expressed as:
\vspace{-3pt}
\begin{eqnarray}
	\mathcal{L}_{semantic}\!=\!\sum_{i=1}^n\omega_i\!\cdot\frac{\left\|F_i(G(x))-F_i(P)\right\|_1}{\left\|F_i(G(x))\!\!-\!\!F_i(\!N\!)\right\|_1\!+\!1e^{-7}},
\end{eqnarray}

where $F_{i}$ refers to extracting the i-th hidden features from the VGG-16 network pre-trained on ImageNet. $n$ refers to n layers that we choose. P refers to the real images of the target domain, and N refers to the real images of the source domain. Here we choose the 1st, 3rd, 5th, 9th and 13th layers. $\omega_i$ are weight coefficients, and we set $\omega_1$ = 1/16, $\omega_2$ =1/8, $\omega_3$ = 1/4, $\omega_4$=1/2, $\omega_5$=1.

Besides, to achieve a constraint of the global texture, we use a special feature space. We can build this feature space in any layer of the network, which consists of the correlations between the different filter responses. These feature correlations are given by the Gram matrix:

\begin{eqnarray}
	M^l_{ij}=\sum_{k}f^l_{ik}f^l_{jk},
\end{eqnarray}

where $M^l_{ij}$ is the inner product between the feature maps $i$ and $j$ in the layer $l$. $k$ refers to the vector length. We employ a feature extraction network to extract the feature maps [c, h, w]. We can obtain the feature maps $i$[ c, h*w]  and the feature maps $j$[ c, h*w]  via flatten and matrix transpose. Gram Matrices is the inner product between the feature maps $i$ and $j$. We then get a set of Gram matrices $\{M^1,M^2,\ldots,M^L\}$ from layers \{1,2,3...L\}\ in the feature extraction network. The Gram matrix $M$ is a quantitative description of latent image features. We expect that the distance $d$ (we use L2 to measure this distance) between the generated image G(x) and negatives $N$ is much greater than the distance between the generated image G(x) and positives $P$:
\vspace{-1pt}
\begin{eqnarray}
	\quad d(M(G(x)),M(P))\ll d(M(G(x)),M(N)),\quad\text{}
\end{eqnarray}
\vspace{-1pt}
Therefore, our proposed style contrastive loss can be expressed as:
\vspace{-1pt}
\begin{eqnarray}
	\begin{aligned}\mathcal{L}_{style}=\text{max}\left\{d(M(G(x)),M(P))\right.\\\left.-d(M(G(x)),M(N))+\alpha,0\right\},\end{aligned}
\end{eqnarray}
\vspace{-1pt}
where $\alpha$ are hyperparameters (similar to the triplet loss \cite{Florian} ), we set it to 0.04 in our experiments. 
Finally, our dual contrastive Regularization is formulated as:
\begin{eqnarray}
	\mathcal{L}_{Dual}=\lambda_1\mathcal{L}_{semantic}+\lambda_2\mathcal{L}_{style},
\end{eqnarray}

where $ \lambda_{1} $  and $\lambda_{2}$ are weight coefficients, and we set $\lambda_{1}$   = 1, $\lambda_{2}$  =0.5.

\begin{table*}[t]
	\centering
	\caption{Quantitative comparison with all baselines.\label{tab1}}
	\centering
			\begin{tabular}{lcccccc}
				\toprule
				\multirow{2}{*}{\textbf{Method}}  & \multicolumn{2}{c}{\textbf{Cat$\rightarrow$Dog}} & \multicolumn{1}{c}{\textbf{Van gogh$\rightarrow$Photo}} & \multicolumn{3}{c}{\textbf{ Horse$\rightarrow$Zebra}} \\ 
				\cmidrule(l){2-3} \cmidrule(l){4-4} \cmidrule(l){5-7}
				& \textbf{FID}$\downarrow$ & \textbf{SWD}$\downarrow$ & \textbf{FID}$\downarrow$ & \textbf{FID}$\downarrow$ & \textbf{SWD}$\downarrow$ & \textbf{sec/iter}$\downarrow$  \\
				\midrule
				{CycleGAN} & 80.5 &  19.5 & 103.0 & 72.2 & 39.1 & 0.40 \\ 
				{CUT} & 76.2 & 12.9  & 96.9 & 45.5 & 31.5 & 0.24\\
				{FastCUT} & 94.0 & 17.6 & 105.3 & 73.4 & 38.2 & 0.15\\
				{FSeSim} & 87.8 & 13.8 & 94.3 & 43.4 & 37.2 & \textbf{0.11}\\ 
				{DCLGAN} & 68.7 & 12.5 & 93.7 & 43.2 & 31.2 & 0.41\\ 
				{QS-Attn} & 72.8  & 12.8 & 92.2 & 41.1 & 30.3 &0.22 \\
				\midrule 
				{SN-DCR} & \textbf{62.7}  & \textbf{12.1}  & \textbf{90.3} & \textbf{33.6} & \textbf{28.4} & 0.23 \\ 
				\bottomrule
			\end{tabular}
		
		\begin{tablenotes}
			\item For the metrics, our algorithm outperforms all the baselines obviously, and our SN-DCR achieves state-of-the-art performance on unpaired I2I translation tasks.
		\end{tablenotes}
	\end{table*}

\subsection{Patch-wise Contrastive Loss}

Following the setup of CUT, we employ a patch-wise contrastive loss to maximize the mutual information between the inputs and outputs. The key idea is to pull the query (patch from the generated image, yellow box in Fig.2) and positives (corresponding from the input image, blue box) together, and push the query and negatives (non-local patches from the input image, red box) away. We employ E to extract features from the input image and the generated image. Then we introduce QS-Attn module to deliberately choose important anchors for patch-wise contrastive loss. The formula can be expressed as:
\begin{small}
\begin{eqnarray}
	\ell=-\log\left[\dfrac{\exp\left(q\cdot k^{+}/\tau\right)}{\exp\left(q\!\cdot\! k^{+}/\tau\right)\!+\!\sum_{n=1}^{N-1}\exp\left(q\!\cdot\! k^{-}/\tau\right)}\right],
\end{eqnarray}
\end{small}
where $q$ refers to the anchor feature (yellow box in Fig.2) from G(x), $k^{+}$ refers to a positive (blue box) and $k^{-}$ refers to (N-1)
negatives (red box). Here $\tau$ indicates a temperature parameter used to measure the distance between query and other samples. The default value is $0.07$ . Note that the positive $k^{+}$ is the corresponding of the anchor feature $q$ in the input image x , and $(N-1)$ negatives are randomly selected in x.

We utilize E to extract features from the generated image, select the L layers of interest from E, and sent it to Q (QS-Attn moudle) to get the features we need. The resulted features can be denoted by $\{\hat{\mathbf{z}}_l\}_\text{L}=\{Q_l(E^l(G(\mathbf{x})))\}_L$, where $E^l$ represents l-th layer we choose. We use qs-attn module to select important anchors in each selected layer , $s\in\left\{1,...,S_l\right\}$ ($S_l$ represent the number of patches selected in each layer). In the same way, the corresponding patches of L layer are obtained from the input image, $\{{\mathbf{z}}_l\}_\text{L}=\{Q_l(E^l(G(\mathbf{x})))\}_L$. We take the corresponding patches obtained from the input image as positives, and the other features as negatives.

The patch-wise contrastive loss can be expressed as:
\begin{small}
\begin{eqnarray}
	\mathcal{L}_{_{patch}}(G,Q,X)=\mathbb{E}_{\boldsymbol{x}^{\sim}X}\sum_{l=1}^{L}\sum_{s=1}^{S_l}\ell\Big(\boldsymbol{\hat{z}}_l^s,\boldsymbol{z}_l^s,\boldsymbol{z}_l^{S\setminus s}\Big), 
\end{eqnarray}
\end{small}
In summary, the overall objective function of SN-DCR can be formulated as:
\begin{eqnarray}
	\mathcal{L}_{Total}=\mathcal{L}_{adv}+\mathcal{L}^{X}_{patch}+\mathcal{L}^{Y}_{patch}+\mathcal{L}_{Dual},
\end{eqnarray}

where $\mathcal{L}^{X}_{patch}$ is the patch-wise contrastive loss defined in Eq.(8), and  $\mathcal{L}^{Y}_{patch}$ is the identity loss, in which the positive $k^+$ and negatives $k^-$ are extracted from 
a real image y from the domain Y, and the anchor q is from G(y). We refer to the CUT settings to add the identity loss. The goal of this identity loss is to prevent generator G from making changes on the target domain images.

	\begin{figure*}[t]
		\centering
		\includegraphics[width=1.0\linewidth]{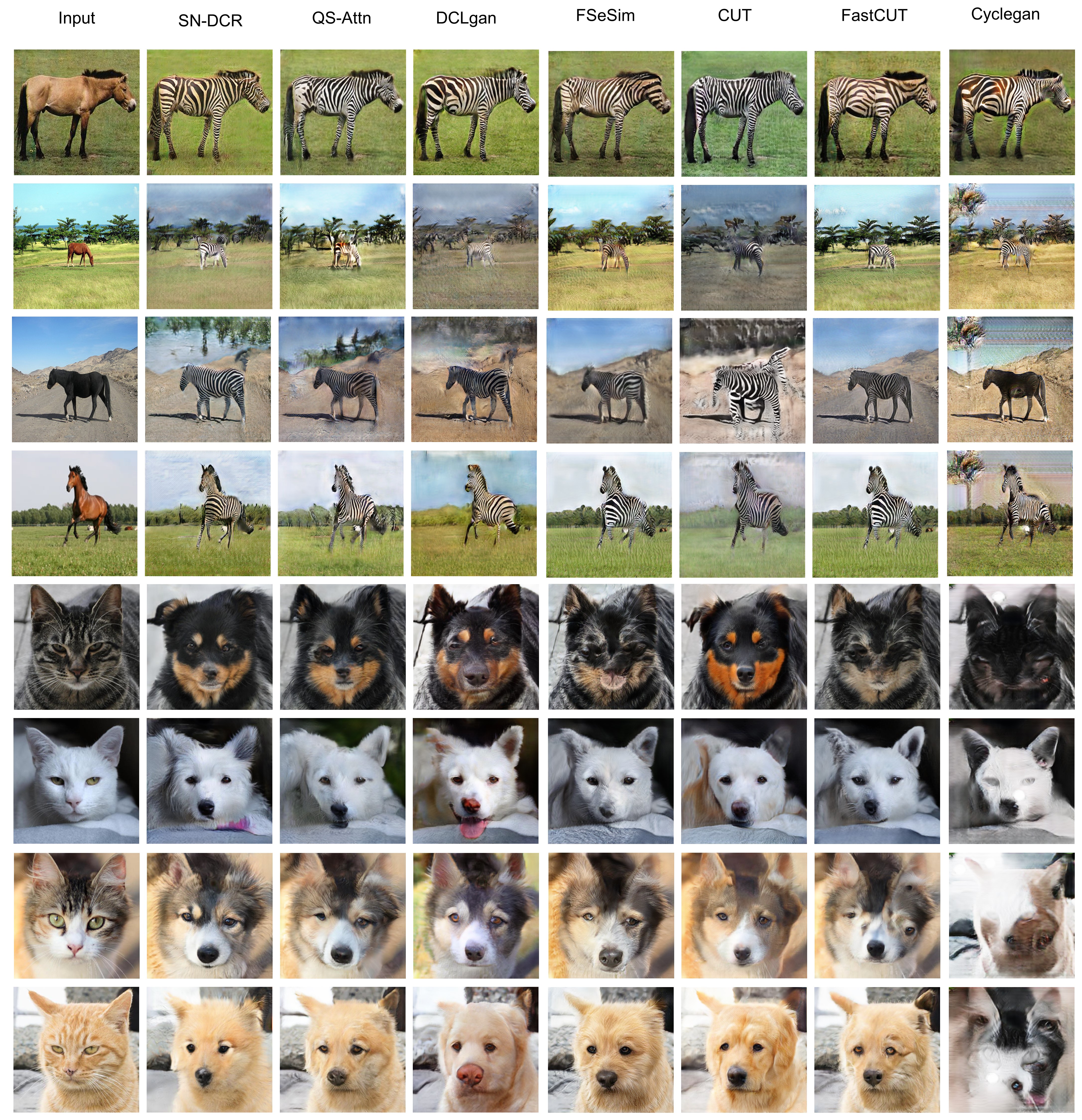}
		\caption{
			Visual results comparison with all baselines on the Horse $\rightarrow$ Zebra and Cat$\rightarrow$ Dog datasets. Compared with all baselines, our SN-DCR not only performs better in terms of global structure and texture, but also generates more real images with more natural detail.
		}
		\label{fig:framework}
	\end{figure*}

	\begin{center}
		\begin{table*}[t]%
			\caption{ Efficiency comparison with all baselines.\label{tab1}}
			\centering
			\begin{tabular}{lccccc}
				\toprule
				&\multicolumn{3}{@{}c@{}}{\textbf{Horse2Zebra}}
				\\\cmidrule{2-5}
				\textbf{Method} & \textbf{Memory}$\downarrow$  & \textbf{Overall Training time}$\downarrow$  & \textbf{FLOPs}$\downarrow$  & \multicolumn{1}{@{}l@{}}{\textbf{Parameters}}$\downarrow$    \\
				\midrule
				{CycleGAN} & 4.7G & 46h &128.26G & 28.286M  \\ 
				{CUT} & 3.9G & 27h &64.13G& 14.703M \\
				{FastCUT} & \textbf{3.4G} & 17h &64.13G & 14.703M \\
				{FSeSim} & 3.8G & \textbf{12h}& 64.13G & \textbf{14.143M} \\ 
				{DCLGAN} & 7.8G  & 44h &128.26G& 29.406M \\ 
				{QS-Attn} & 5.5G & 24h &64.13G & 14.703M  \\
				\midrule 
				{SN-DCR} & 5.5G & 26h & \textbf{42.38G} &14.679M  \\ 
				\bottomrule
			\end{tabular}
			\begin{tablenotes}
				\item Our proposed SN-DCR demonstrates competitive performance in the realm of image-to-image translation, notwithstanding its suboptimal outcomes in terms of computational speed, parameter efficiency, and memory consumption. Despite these limitations in specific computational metrics, SN-DCR distinguishes itself by exhibiting notable efficacy in the image translation task. This underscores its merit as a viable alternative that strikes a judicious balance between considerations of computational resources and translation performance, thereby constituting a valuable contribution to the field. Moreover, we conduct a comparative analysis of FLOPS for generator of each methods, and our approach has demonstrated superior performance, which further substantiates the advantage of SN-DCR.
			\end{tablenotes}
		\end{table*}
	\end{center}

		\begin{center}
		\begin{table*}[t]%
			\caption{ Quantitative comparison with all baselines on Cityscapes dataset.\label{tab1}}
			\centering
			\begin{tabular}{lcccccc}
				\toprule
				&\multicolumn{5}{@{}c@{}}{\textbf{CityScapes}}
				\\\cmidrule{2-6}
				\textbf{Method} & \textbf{FID}$\downarrow$  & \textbf{mAP}$\uparrow$  & \multicolumn{1}{@{}l@{}}{\textbf{pixAcc}}$\uparrow$   & \textbf{classAcc}$\uparrow$  & \textbf{SSIM}$\uparrow$    \\
				\midrule
				{CycleGAN} & 76.3 & 20.4 & 55.9 & 25.4 & 0.3601 \\ 
				{CUT} & 56.4 & 24.7 & 68.8 & 30.7 & 0.4474\\
				{FastCUT} & 68.8 & 19.1 & 59.9 & 24.3 & 0.4265\\
				{FSeSim} & 54.3 & 22.1 & 69.4 & 27.8 & 0.4367\\ 
				{DCLGAN} & 49.4  & 22.9 & 76.9 & 29.6 & 0.4486\\ 
				{QS-Attn} & 53.5 & 25.5 & \textbf{79.9} & 31.2 & 0.4573 \\
				\midrule 
				{SN-DCR} & \textbf{46.6} & \textbf{27.9} & 74.3 & \textbf{35.4} & \textbf{0.4638} \\ 
				\bottomrule
			\end{tabular}
			\begin{tablenotes}
				\item For Cityscapes dataset, we use the pretrained semantic segmentation network DRN, and compute mean average precision (mAP), pixel-wise accuracy (pixAcc), and average class accuracy (classAcc), showing the semantic interpretability of the generated images. Unlike other datasets, CityScapes does have corresponding labels. So we conduct an evaluation between ground truth and the generated images via structural similarity (SSIM) \cite{DBLP:journals/tip/WangBSS04}  metrics to evaluate the structure consistency.
			\end{tablenotes}
		\end{table*}
	\end{center}

		\begin{figure*}[t]
		\centering
		\includegraphics[width=1.0\linewidth]{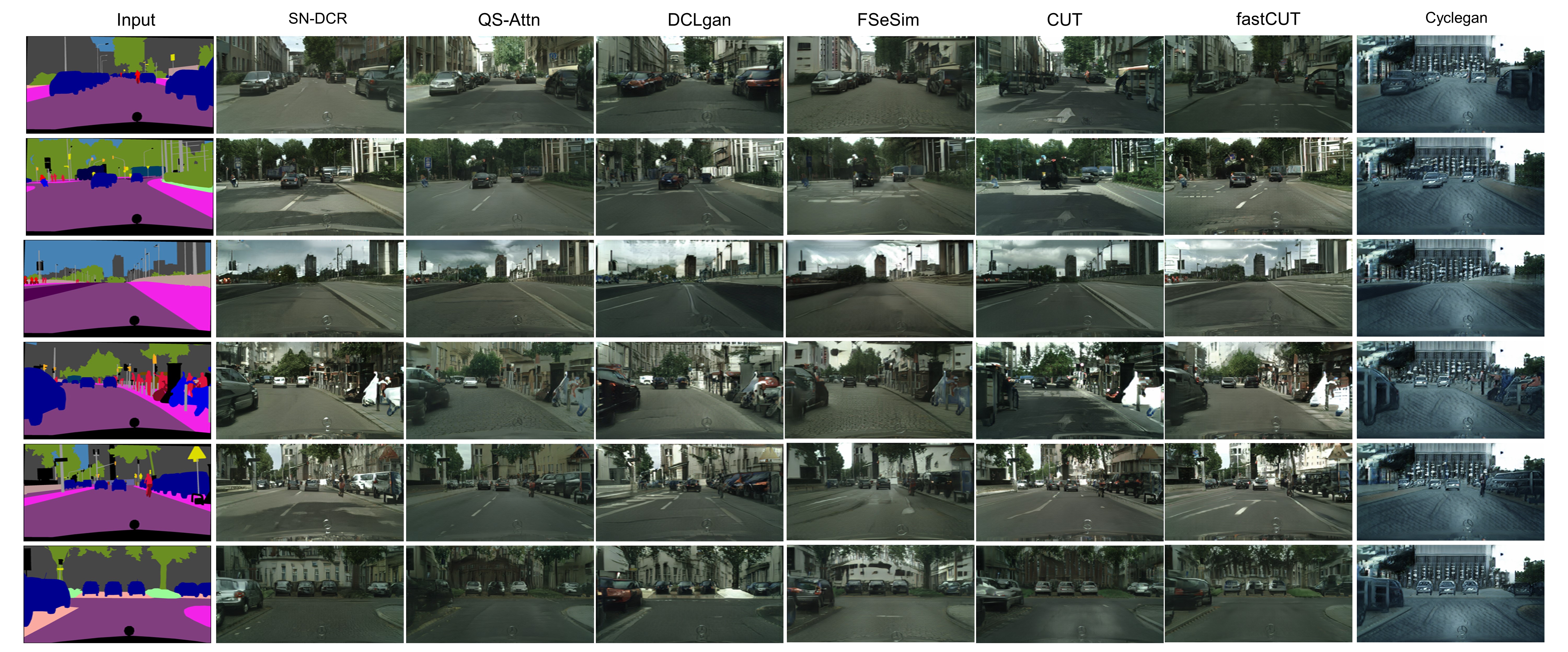}
		\caption{
			Visual results comparison with all baselines on the CityScapes dataset. Our proposed SN-DCR show visual satisfactory results. SN-DCR can clearly generate a wide variety of cars with natural details, and performs better in terms of global structure and texture.
		}
		\label{fig:framework}
	\end{figure*}

		\begin{figure*}[t]
		\centering
		\includegraphics[width=1.0\linewidth]{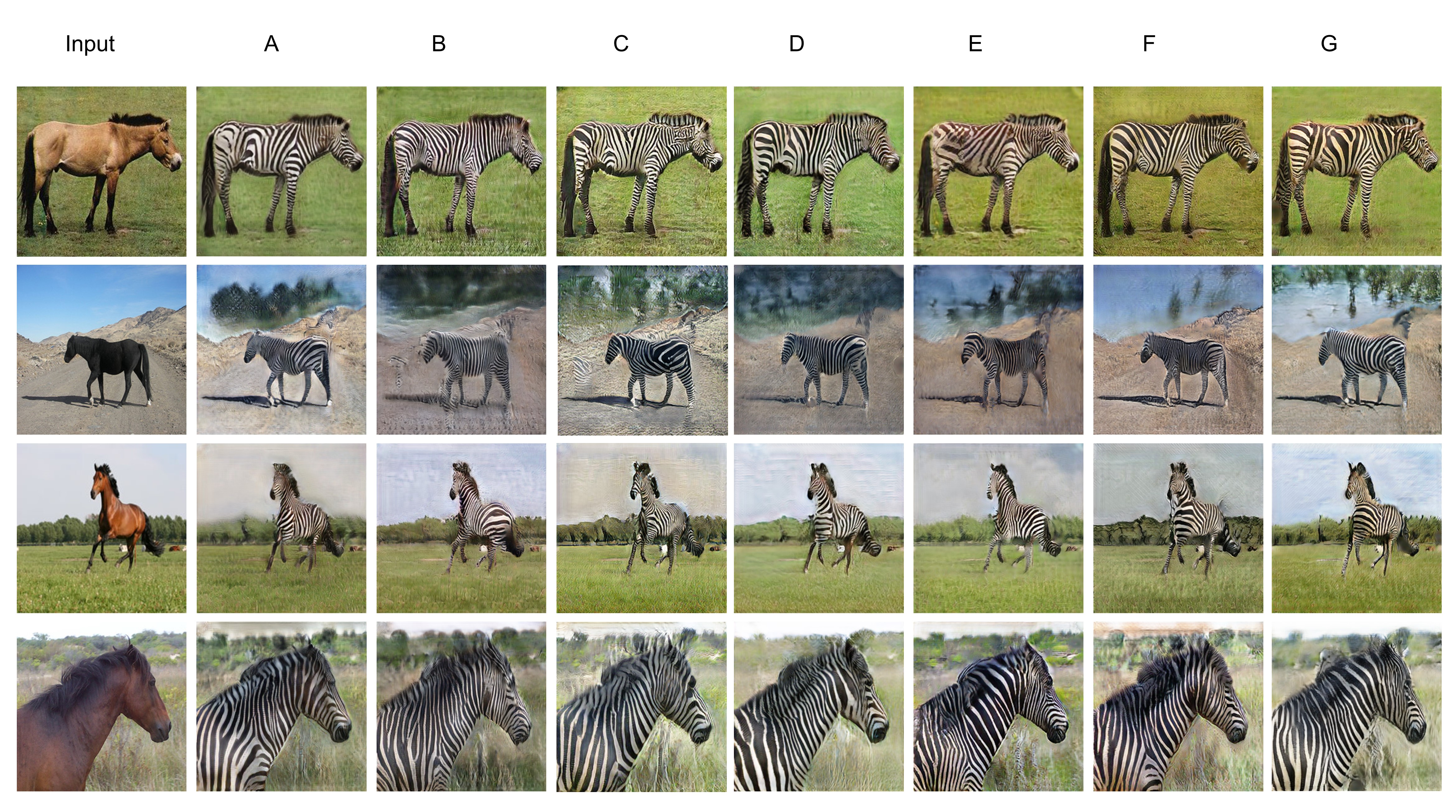}
		\caption{
			Visual results for ablation study on Horse$\rightarrow$ Zebra dataset. The leftmost column are input images. Model G is our proposed SN-DCR.
		}
		\label{fig:framework}
		\vspace{3pt}
		
	\end{figure*}

	\begin{center}
		\begin{table*}[t]%
			\caption{ Ablation study on Horse $\rightarrow$ Zebra dataset. ( hyperparameter $ \lambda_{1} $ and $ \lambda_{2} $ on DCR)\label{tab1}}
			\centering
			\begin{tabular}{lcccccc}
				\toprule
				
				\textbf{ } & \textbf{$ \lambda_{1} $=0.1}  & \textbf{$ \lambda_{1} $=0.5 } & \multicolumn{1}{@{}l@{}}{\textbf{$ \lambda_{1} $=1}}   & \textbf{$ \lambda_{1} $=5}  & \textbf{$ \lambda_{1} $=10}    \\
				\midrule
				{\textbf{$ \lambda_{2} $=0.1}} & 38.1 & 37.9 & 36.5 & 37.4 & 38.4 \\ 
				{\textbf{$ \lambda_{2} $=0.5}} & 38.3 & 34.1 & \textbf{33.6} & 36.7 & 37.9\\
				{\textbf{$ \lambda_{2} $=1}} & 39.2 & 35.5 & 33.9 & 36.2 & 38.9\\
				{\textbf{$ \lambda_{2} $=5}} & 40.3 & 37.8 & 35.4 & 38.6 & 40.9\\ 
				{\textbf{$ \lambda_{2} $=10}} & 41.1  & 38.4 & 36.2 & 39.5 & 40.7\\ 
				
				\bottomrule
			\end{tabular}
			\begin{tablenotes}
				\item  We  use  FID  to evaluate ablation study.
				
			\end{tablenotes}
		\vspace{-5pt}
		\end{table*}
	\end{center}

	\begin{table*}[t]
		\centering
		\caption{Ablation study on Horse $\rightarrow$ Zebra dataset.\label{tab1}}
		\centering
				\begin{tabular}{lcccccc}
					\toprule
					\textbf{ } &\multicolumn{5}{c}{\textbf{ Configuration}}\\
					\cmidrule(l){2-6}
					\multirow{2}{*}{\textbf{Method}}  & \multicolumn{2}{c}{\textbf{Generator}} & \multirow{2}{*}{\textbf{QS-Attn}}  & \multicolumn{2}{c}{\textbf{ DCR}} &\multirow{2}{*}{\textbf{FID}}  \\ 
					\cmidrule(l){2-3} \cmidrule(l){5-6}
					& \textbf{FCA} & \textbf{SN}& \textbf{ } & \textbf{semantic} & \textbf{style}   \\
					\midrule
					{CUT} &   &    &   &   &   & 45.5\\
					\midrule 
					{A} &  $\times$ &  $\times$ &  $\times$ & $\times$  &  $\checkmark$&  37.4\\
					{B} & $\times$ & $\times$ & $\times$ & $\checkmark$ & $\times$&38.7 \\
					{C} & $\times$& $\times$ & $\times$ &  $\checkmark$ &  $\checkmark$ & 35.5\\ 
					\midrule
					{D} & $\times$ &$\times$ &  $\checkmark$ &  $\checkmark$ &  $\checkmark$ &34.5\\ 
					{E} &  $\checkmark$  &  $\checkmark$ &  $\checkmark$ & $\times$ & $\times$ &38.9\\
					
					{F} & $\times$ &  $\checkmark$  &  $\checkmark$ &  $\checkmark$ & $\checkmark$ & 34.1\\ 
					\midrule 
					{G} &  $\checkmark$  &  $\checkmark$ &  $\checkmark$ &  $\checkmark$ &  $\checkmark$ &\textbf{33.6} \\
					\bottomrule
				\end{tabular}
				\begin{tablenotes}
					\item DCR refers to dual contrastive regularization. semantic means semantically contrastive loss, and style means style contrastive loss. FCA refers to the Frequency Channel Attention Network. SN refers to the spectral normalized convolutional network. QS-Attn refers to QS-Attn moudle.
					
				\end{tablenotes}
			\vspace{-0.1cm}
			\vspace{10pt}
			\vspace{-0.3cm}
			\label{tab:tab1}
		\end{table*}

\section{ EXPERIMENTS}\label{sec4}

\subsection{Experimental Settings}

\noindent\textbf{Datasets.} SN-DCR is trained and evaluated on Horse$\rightarrow$ Zebra, Cat$\rightarrow$Dog , Van Gogh $\rightarrow$ Photo and CityScapes datasets. Horse$\rightarrow$Zebra is provided in \cite{Benson1992} , which contains 1,067 and 1,334 training images for horse and zebra, respectively. We use 120 horse images as the test images on Horse $\rightarrow$ Zebra. Cat $\rightarrow$ Dog is from \cite{Yunjey}, which consists of 5,153 and 4,739 training images for cat and dog, respectively. We used 500 images of cats as test images on Cat $\rightarrow$ Dog. Van Gogh $\rightarrow$ Photo is a dataset of 400 Van Gogh paintings and 6287 photos extracted from \cite{Yunjey}. We used 400 Van Gogh images as test images. Cityscapes contains street scenes from German cities, with 2,975 training images and 500 test images.

\noindent\textbf{Training details.} The implementation of SN-DCR is mainly based on CUT. We use a ResNet-based generator and a PatchGAN discriminator. Different from CUT, we introduce qs-attn module in patch-wise contrastive loss, and introduce SN and FCA module in design of the generator. Our proposed dual contrastive regularization employs VGG16 to extract features. we use the Adam optimizer, $\beta_{1}= 0.5$ and $\beta_{2}= 0.999$. The batch size we used is 1, and all training images are loaded into 286 * 286, then cut to 256 * 256 blocks. SN-DCR trains 400 epochs on each dataset, the learning rate is 0.0002, and the learning rate starts to decay linearly to 0 after 200 epochs.

\noindent\textbf{Evaluation metrics.} 
We primarily utilize Frechet Inception Distance (FID) \cite{Martin} and Sliced Wasserstein Distance (SWD) \cite{Alfred} as key evaluation metrics for assessing the performance of SN-DCR. Both FID and SWD are well-established metrics with a robust correlation to human perception. They measure the distance between two distributions, namely the distributions of real and generated images. A lower FID and SWD signify a closer resemblance between the generated and real images. For Cityscapes dataset, we use the pretrained semantic segmentation network DRN \cite{Fisher} , and compute mean average precision (mAP), pixel-wise accuracy (pixAcc), and average class accuracy (classAcc), showing the semantic interpretability of the generated images.

\subsection{Comparison with other Methods}
\vspace{3pt}
Table 1 shows the quantitative results of our proposed SN-DCR compared with all baselines on three datasets, including QS-Attn \cite{Xueqi}, DCLgan  \cite{Margolin2003} , FseSim \cite{Chuanxia}, CUT \cite{Dukowicz1984}, FastCUT \cite{Dukowicz1984} and Cyclegan \cite{Benson1992}. We mainly use FID and SWD scores as our quantitative metrics. For the metrics, our algorithm outperforms all the baselines obviously, and our SN-DCR achieves state-of-the-art performance on unpaired I2I translation tasks. As illustrated in the Table 2, SN-DCR demonstrates superiority over DCLgan and Cyclegan concerning both performance and computational parameters. Despite having computational parameters comparable to QS-Attn and CUT, SN-DCR outperforms them in terms of performance. Although SN-DCR falls slightly short of FastCUT and FSeSim in computational parameters, its performance significantly surpasses that of FastCUT and FSeSim. Despite these limitations in specific computational metrics, SN-DCR distinguishes itself by exhibiting notable efficacy in the image translation task. Fig.1 shows the visual results comparison with all baselines on the Van Gogh$\rightarrow$Photo dataset. Compared with other methods, our SN-DCR is able to preserve the global structure information and generate the photos with more natural details, proving that our DCR is effective to improve the global information. To better illustrate the superiority of our SN-DCR, we display some visual results in Fig.5. We randomly pick four samples on the Horse $\rightarrow$ Zebra and Cat $\rightarrow$ Dog datasets. QS-Attn, DCLGAN and FseSim fail to preserve details and textures of the generated image. CUT, FastCUT, Cyclegan generate unsatisfactory content. Compared with all baselines, our SN-DCR not only performs better in terms of global structure and texture, but also generates more real images with more natural details.
Fig 6 shows the visual results comparison with all baselines on the CityScapes. Compared with other methods, our SN-DCR is able to clearly generate a wide variety of cars with natural details, and performs better in terms of global structure and texture, proving that our DCR is able to enhance consistency of the global texture. Moreover, our SN-DCR can produce more real images with more color details. Table 2 shows the quantitative results of our proposed SN-DCR compared with all baselines on the CityScapes datasets. Obviously, our algorithm outperforms all the baselines. For structural similarity (SSIM) metrics, our proposed SN-DCR performs better than other methods, proving that our DCR is effective to enhance the information of global structure.

\subsection{Ablation Study}

In comparison experiments, SN-DCR shows better performance than all baseline methods. In SN-DCR, we apply the FCA and SN module in design of the generator, QS-Attn module in patch-wise contrastive learning, and dual contrastive regularization.

For effect of the weights( hyperparameter $ \lambda_{1} $ and $ \lambda_{2} $ ) in DCR, ablation experiments are performed on the Horse $\rightarrow$ Zebra dataset to ensure their optimum values, as shown in Table 3. To evaluate the measures separately, we mainly conduct the ablation study on the Horse $\rightarrow$ Zebra dataset. Our baseline is CUT. Visual results and metrics for ablation study are listed in Fig 7 and Table 4.  

The metric of models A and B  performs better than CUT, reflecting that our proposed global contrastive loss functions are useful for I2I translation tasks. Model C outperforms A and B, indicating the effectiveness of our dual setting. Model D performs better than C, indicating that the addition of QS-Attn module is beneficial to the model's performance. Model E without DCR to exhibit the different influence between global constraints and local constraints. Model E outperforms QS-Attn, proving that our designed generator is effective. Model F outperforms D, indicating the effectiveness of the SN ResBlock in our generator. Finally, Model G, which combines all the measures, achieves the best performance on I2I translation tasks, validating the efficacy of our overall approach. 

\section{ CONCLUSION}\label{sec5}

In this paper, we propose a new unpaired I2I translation framework based on dual contrastive regularization and spectral normalization, namely SN-DCR. To achieve a global constraint of structure and texture in an unpaired manner, we formulate two new global contrastive loss functions to supplement the patch-wise contrastive loss, called dual contrastive regularization (DCR). To alleviate mode collapse and convergence difficulties, we employ the spectral normalized convolutional network in the design of our generator. Moreover, to further boost the feature representation ability of our model, Frequency Channel Attention Network is introduced to further boost the feature representation ability of our generator.  In ablation study, we have proved the effectiveness of each component of our method. Our design can better deal with various tasks of I2I translation. In comparison experiments, SN-DCR shows better performance than all baseline methods in terms of global structure and texture. In multiple datasets, quantitative and visual results demonstrate that SN-DCR achieves the best results. Furthermore, we highlight that the principal advantage of SN-DCR lies in its remarkable capability to generate highly realistic images, particularly excelling in the generation of textures and natural details. Nevertheless, it is noteworthy that our approach comes with certain constraints, chief among which are less than optimal performance in terms of both training costs and memory consumption. Finally, we believe that our work will initiate further research on unpaired image-to-image translation.

\noindent\textbf{Acknowledgements} This work was supported in part by the National Natural Science Foundation of China under Grant No. 62276138 and Grant No. 61876087.

\noindent\textbf{Data availability statement} The datasets generated during and/or analysed during the current study are available from the corresponding
author on reasonable request.

\section*{ Declaration}
\noindent\textbf{Conflict of interest} The authors declare that they have no conflict of interest.

\bibliographystyle{sn-basic.bst} 
\bibliography{sn-bibliography.bib}



\end{document}